\documentclass{article}

\usepackage{iclr2027_conference,times}

\usepackage[utf8]{inputenc}
\usepackage[T1]{fontenc}
\usepackage{graphicx}
\usepackage{booktabs}
\usepackage{multirow}
\usepackage{amsfonts}
\usepackage{amssymb}
\usepackage{nicefrac}
\usepackage{microtype}
\usepackage{xcolor}
\usepackage{amsmath}
\usepackage{eccvabbrv}
\usepackage{subcaption}
\usepackage{hyperref}
\usepackage{url}
\usepackage[capitalize]{cleveref}
\usepackage{wrapfig}

\usepackage{amsmath}

\newcommand{\vect}[1]{\boldsymbol{\mathbf{#1}}}

\newcommand{\method}{PiHand}

\crefname{equation}{Eq.}{Eqs.}
\Crefname{equation}{Equation}{Equations}
\crefname{figure}{Fig.}{Figs.}
\Crefname{figure}{Figure}{Figures}
\crefname{section}{Sec.}{Secs.}
\Crefname{section}{Section}{Sections}
\crefname{table}{Tab.}{Tabs.}
\Crefname{table}{Table}{Tables}

\title{\method{}: Latent Hand–Object Interaction Priors for Physics-Based Policy Learning}

\author{\makebox[\dimexpr\textwidth-2\tabcolsep\relax][s]{Niklas Berndt$^{1,2,3}$\hfill Xianghui Xie$^{1,2,3}$\hfill Enes Duran$^{1,2}$\hfill Yixiao Wang$^{4}$\hfill Gerard Pons-Moll$^{1,2,3}$}\\ \parbox[t]{\dimexpr\textwidth-2\tabcolsep\relax}{\normalfont \mbox{$^{1}$University of Tübingen, Germany}\quad \mbox{$^{2}$Tübingen AI Center, Germany}\quad \mbox{$^{3}$Max Planck Institute for Informatics, Saarland Informatics Campus, Germany}\quad \mbox{$^{4}$University of California, Berkeley, USA}}\\[0.6em] \normalfont \url{https://virtualhumans.mpi-inf.mpg.de/PiHand/}}

\iclrfinalcopy

\begin{document}

\maketitle

\begin{figure}[ht]
    \centering
    \includegraphics[width=\linewidth]{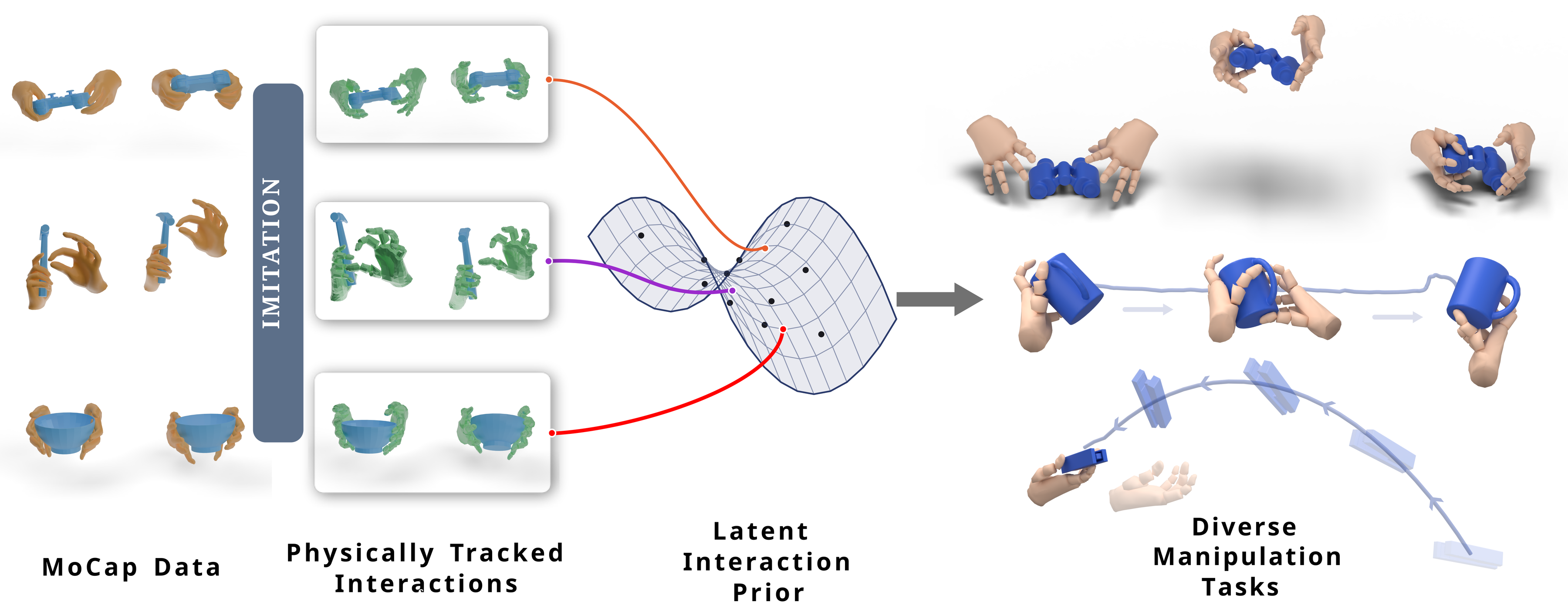}
    \caption{We propose \method{}, a method to generate physically plausible hand-object interaction motions by learning physically grounded interaction priors from mocap references. Our single interaction prior transfers to diverse dexterous manipulation tasks.}
    \label{fig:teaser}
\end{figure}
\begin{abstract}
Synthesizing physically grounded dexterous hand--object interactions is important for VR/AR, gaming, and robotics, yet remains challenging because subtle contact dynamics, diverse object geometries, and target trajectories must be jointly respected while maintaining natural grasps. Existing methods either track mocap trajectories with limited transfer to novel objects, or rely on hand-only generative priors that fail to capture coupled hand--object motion, sacrificing contact fidelity and sample efficiency. To tackle these challenges, we introduce \textbf{\method{}}, a \textbf{P}hysically grounded \textbf{I}nteraction-aware \textbf{Hand} motion synthesis framework. \method{} first trains expert policies that track mocap demonstrations in simulation, distills them into a unified hand--object interaction prior using a goal-conditioned VAE, and finally solve diverse dexterous manipulation tasks through bounded residual exploration in its latent space. With only a single reward and no complex curriculum training, this yields general tabletop manipulation policies that generalize to unseen objects and trajectories on GRAB and DexYCB. The same prior transfers to object catching and bimanual handover, two tasks that are rarely seen in its training data.
Across all tasks, \method{} consistently outperforms baselines in all metrics. We will release the processed data and model checkpoints.
\end{abstract}

\section{Introduction}
\label{sec:intro}
Generating realistic hand-object interactions is a foundational challenge in computer vision and graphics, with implications for VR/AR, gaming and robotics. Several kinematic approaches \citep{cha2024text2hoi, Muchen_LatentHOI, zhang2025bimart} have made significant strides in generating such interactions from visual or textual inputs, yet they frequently suffer from noticeable physical implausibilities which make them unfit for deployment in simulated environments or on real-world robots. To bridge this gap, the robotics and physics-based animation communities have focused on learning generative control policies to synthesize the complex motions required to stably grasp and manipulate objects based on their geometry and target trajectory inside simulation. Yet this remains challenging because contact dynamics are subtle and discrete, objects vary widely in geometry and affordances, and the hand must continuously adapt its grasp and contact configuration to remain both natural-looking and physically correct.

Physics-based tracking controllers imitate mocap sequences~\citep{li2025maniptrans,liu2025dextrack} with strong realism and sim-to-real transfer, but remain reconstructive: their reliance on dense references prevents synthesis or adaptation to new objects without corresponding mocap. Conversely, generative RL methods such as GraspXL~\citep{Zhang2025GraspXL} generalize to unseen objects. But such dexterous-hand exploration in action space requires engineered rewards and curricula, remains sample-inefficient, and often yields unnatural motion. PULSE~\citep{Luo2024PULSE} learns reusable latent humanoid control spaces, which Omnigrasp~\citep{luo2024ominigrasp} applies to object-centric tasks. Yet these human-only priors do not model object interactions, forcing policies to discover contacts in a latent space unstructured for them and limiting scalability to complex, dynamic tasks.

In this work, we demonstrate that an \textit{interaction prior} consistently outperforms both action-space exploration and hand-only priors in sample efficiency, spatial accuracy and tracking robustness. We propose \textbf{\method{}}, a \textbf{P}hysically grounded \textbf{I}nteraction-aware \textbf{Hand} motion synthesis framework that efficiently generates high-quality hand-object interactions, generalizes to new objects and transfers to unseen dexterous tasks. Instead of relying on brittle reward engineering or ``blind'' motion priors, our core insight is to embed the interaction itself into the foundation of the latent space. We argue that such an ``interaction-first'' approach is essential to capture the subtle, contact-rich patterns of human grasping and to construct a well-behaved latent manifold that enables efficient synthesis across novel objects, trajectories and tasks. Specifically, we propose a three-stage pipeline: we first train specialized expert policies in a physics simulator to track hand--object interactions from mocap data, then distill these experts into an object- and contact-aware latent interaction prior, and finally solve new tasks by residual RL exploration within this latent space. 

Experiments show that our imitation experts improve success rate (SR) over the state-of-the-art tracking method from 69.6\% to 93.3\%, providing accurate data for distillation. Using only a simple object-tracking reward, our interaction prior enables tabletop policies to manipulate unseen objects along unseen trajectories with 14.0 and 16.9 percentage points higher SR on GRAB~\citep{GRAB:2020} and DexYCB~\citep{dexycb:cvpr2021}, respectively. The same prior transfers to object catching absent from its training data and sparsely observed bimanual handover, achieving $2.0\times$ and $4.3\times$ the strongest-baseline SR, respectively.

In summary, our main contributions are:

\begin{itemize}
    \item \textbf{Data-Driven Interaction Prior for Efficient RL:} We introduce a data-driven \emph{hand-object interaction prior} that allows efficient RL policy learning for dexterous manipulation, bridging mocap-based human priors with physics-based reinforcement learning.

    \item \textbf{High-Fidelity Interaction Experts:} We learn physically accurate interaction experts to track reference mocap demonstrations for the GRAB dataset, providing a valuable resource for the community and a foundation for learning from human demonstrations.

    \item \textbf{Generalizable and Reusable Interaction Synthesis:} We show that our learned single interaction prior supports unseen objects and trajectories on GRAB and DexYCB, transfers to rare bimanual handovers and unseen object-catching tasks, and consistently outperforms baselines in success rate and spatial controllability.
\end{itemize}

\section{Related Work}

\noindent\textbf{Human and interaction motion imitation.} 
Physics-based motion imitation typically learns reference-conditioned tracking controllers from human motion. Starting from example-guided RL (DeepMimic~\citep{peng2018deepmimic}) and adversarial motion priors/skill embeddings (AMP~\citep{peng2021amp}, ASE~\citep{peng2022ase}), recent general motion tracking systems scale to diverse whole-body behaviors and hardware transfer, e.g., PHC~\citep{Luo2023PHC}, HOVER~\citep{He2025HOVER}, ExBody2~\citep{ji2024exbody2}, GMT~\citep{chen2025gmt}, KungfuBot~\citep{xie2025kungfubot}, and camera/teleop-driven imitators such as H2O~\citep{he2024learning} and TWIST~\citep{ze2025twist}. More recently, VideoMimic~\citep{videomimic} imitates human motion from video and transfers it to a real robot.

Extending imitation to human–object interaction (HOI) requires contact-aware objectives and diverse interaction data: PhysHOI~\citep{Wang2023PhysHOI} introduces contact-graph rewards; SkillMimic~\citep{Wang2025SkillMimic} and SkillMimic-V2~\citep{yu2025skillmimicV2} learn reusable interaction skills from demonstrations and InterMimic~\citep{Xu2025InterMimic} scales to varied objects via teacher–student distillation. Recent loco-manipulation frameworks further build on these trackers with residual refinement or generative priors (MaskedManipulator~\citep{Tessler2025MaskedManipulator}, ResMimic~\citep{zhao2025resmimicgeneralmotiontracking}, DreamControl~\citep{kalaria2025dreamcontrol}, VisualMimic~\citep{yin2025visualmimic}, CoorDex~\citep{li2026coordexcoordinatingbodyhand}).

For hand–object interaction, DexTrack~\citep{liu2025dextrack}, ManipTrans~\citep{li2025maniptrans}, AdaDexTrack~\citep{Adalibieke_2026_CVPR}, HOT~\citep{Wang2025SyntheticHOTracking}, and DexMachina~\citep{mandi2025dexmachinafunctionalretargetingbimanual} learn dexterous tracking control from human references, while DexMimicGen~\citep{jiang2025dexmimicgen} and video-based pipelines (DexMV~\citep{qin2021dexmv}, ViViDex~\citep{chen2024vividex}) expand demonstration coverage; classic demo-augmented RL (DAPG~\citep{rajeswaran2017dapg}) underpins many of these systems. In contrast to predominantly reference-following controllers (or guidance-based synthesis), our method additionally supports generation of novel hand-object motions beyond tracking.

\noindent\textbf{Interaction motion generation.} 
Physics-based approaches often learn task-specific policies for basketball~\citep{Wang2025SkillMimic,yu2025skillmimicV2}, tennis~\citep{zhang2023vid2player3d}, bulky-object manipulation~\citep{zheng2025bulkyobjectshumanoid}, or in-hand rotation~\citep{liu2025dexndm}. Such pipelines require careful reward shaping and curricula and do not readily scale across tasks. To improve reuse and data efficiency, prior work incorporates human-motion representations, ranging from imitation-RL controllers such as DeepMimic~\citep{peng2018deepmimic}, AMP~\citep{peng2021amp}, and PHC~\citep{Luo2023PHC} to reusable skill priors such as ASE~\citep{peng2022ase}. PULSE~\citep{Luo2024PULSE} further distills motion-tracking behavior into a conditional latent representation for downstream latent-space RL. Its overall recipe is the closest precursor to PiHand. We follow this recipe but replace PULSE's human-only whole-body representation with a dexterous hand--object prior that jointly models hand state, object geometry and trajectory, and contact dynamics.

Building on these priors, recent works move toward modelling object manipulations: Omnigrasp leverages pretrained humanoid motion representations to learn grasp-and-carry across more than 1,000 objects~\citep{luo2024ominigrasp}, GraspXL scales goal-conditioned grasp-motion synthesis to massive unseen object sets~\citep{Zhang2025GraspXL}, and MaskedMimic reframes physics-based control as masked motion inpainting conditioned on sparse cues (e.g., keyframes/text/object goals)~\citep{tessler2024maskedmimic}. TokenHSI further unifies diverse physical human–scene skills via task tokenization in a single transformer policy~\citep{pan2025tokenhsi}, but still largely relies on human-only priors. Only a few works directly learn interaction priors: concurrent work InterPrior~\citep{xu2026interprior} distills full-reference HOI experts into a goal-conditioned variational controller and RL-finetunes for generalization, complementary to kinematic HOI priors that explicitly model human–object coupling (DecHOI~\citep{jung2025dechoi}, CoopDiff~\citep{lin2025coopdiff}); MaskedManipulator similarly distills a tracking expert into a goal-conditioned generative loco-manipulation policy over coupled body–object objectives~\citep{Tessler2025MaskedManipulator}. Orthogonal to these, some kinematic methods~\citep{li2023omomo, li2024chois, zhang2022couch, zhang2024force, Zhou_2024_GEARS, Taheri:CVPR:2022:GOAL, Wu_2025_human_instruction, peng2023hoi} learn interaction priors by design but their physical plausibility is limited without simulation. Overall, many systems still rely on human-only priors or task-specific rewards rather than a learned interaction prior, motivating our interaction-first approach for more efficient learning and accurate manipulation.

\section{Method}
\label{sec:method}
\begin{figure}[t]
    \centering
    \includegraphics[width=1.0\linewidth]{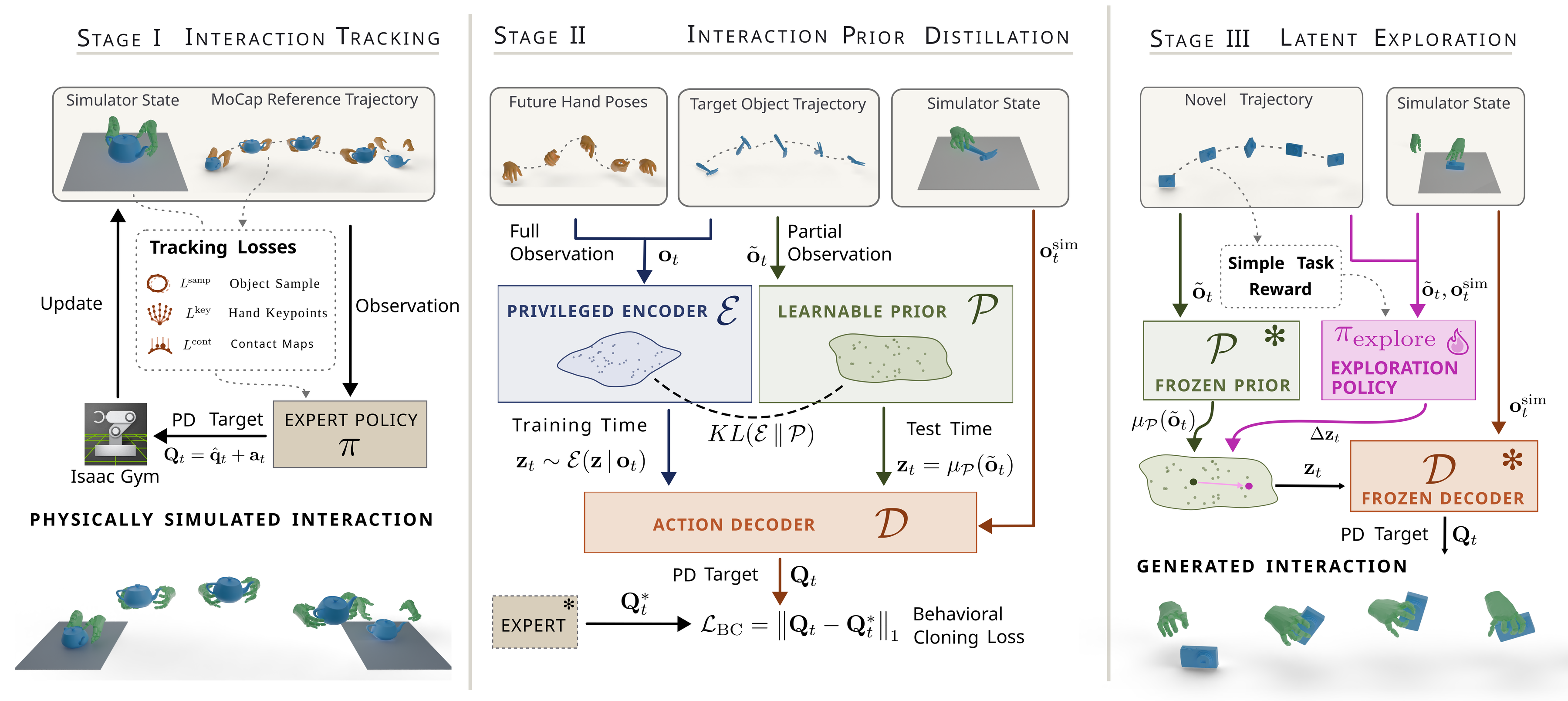}
    \caption{\textbf{Method Overview}. In Stage I we train expert policies that can track mocap hand-object interactions in a physically simulated environment. In Stage II we then distill the expert knowledge into a VAE-like student network via behavior cloning. The VAE embeds hand-object interactions into a latent space which we leverage in Stage III for efficient sampling and exploration for object trajectory guided motion generation.}
    \label{fig:method}
\end{figure}

Our method synthesizes physically plausible hand--object interactions conditioned only on a target object trajectory. The key idea is to construct a latent space that embeds data-driven priors of physically simulated hand-object interaction motions. As illustrated in \Cref{fig:method}, our pipeline consists of three stages.
In Stage I we learn a set of per-object expert policies that convert noisy mocap demonstrations into physically grounded hand-object interaction motions. In Stage II, we distill the expert's behaviors into a latent interaction prior. Finally, in Stage III we explore this latent space, allowing the model to efficiently generate new interactions and solve novel tasks.

\noindent\textbf{Preliminaries.}
Our simulation environment contains two PD-actuated hands, an unactuated rigid object, and a static table. Our \textbf{hand state}  $\vect q=(\vect q^L,\vect q^R)$ consists of the left hand $\vect q^L$ and the right hand $\vect q^{R}$ where each hand $\vect{q}^h=(\vect{P}_w^h, \vect{\theta}^h), h \in \{L, R\}$ is represented by the wrist pose $\vect P_w^h\in SE(3)$ and internal finger joint angles $\vect\theta^h \in \mathbb{R}^{45}$. Each hand has a total of $n_D=51$ DoF, $n_B = 16$ collision bodies, and we additionally use $n_K=21$ kinematic keypoints per hand, including the wrist, internal joints, and fingertips, to compute rewards on hand pose and contact. The passive object is represented by its pose $\vect P_o\in SE(3)$ and linear and rotational velocities $\dot{\vect P}_o$, and evolves only under gravity and contact forces. We use $\hat{\cdot}$ to denote reference quantities from mocap throughout this section. We treat $\vect P=(\vect R,\vect t)\in SE(3)$ as acting on points by
$\vect P\vect x = \vect R\vect x + \vect t$.

\subsection{Stage I: Physics-Grounded Hand--Object Interaction Tracking} \label{subsec:tracking_v2}
Mocap demonstrations record kinematics but not dynamics: contacts are frequently inconsistent and forces are absent altogether. In Stage I we aim to produce physically valid interaction data by training expert tracking policies via PPO~\citep{Engstrom2017ProximalPO} to imitate mocap demonstrations in a simulator. The resulting experts are then used in Stage II to learn a latent interaction prior for motion generation.

Because the mocap reference is already a strong prior on the desired motion, the policy need only correct it rather than synthesize control from scratch. We therefore parameterize the PD target at time $t$ as the reference-anchored pose $\vect Q_t = \hat{\vect q}_t + \vect a_t$, where $\vect a_t \in \mathbb{R}^{2n_D}$ is a learned residual correction in pose space. Predicting $\vect a_t$ directly, however, allows the policy to make
large frame-to-frame jumps in the PD target, which produces jittery
control and unstable contacts. Instead, $\pi$ outputs a per-step
\emph{increment} that we integrate,
$\vect a_t = \vect a_{t-1} + \pi(\vect o_t)$, with $\vect a_0 = \vect 0$,
so that temporal smoothness is built into the action parameterization.
We further constrain $\vect a_t$ to per-joint deviation bounds around
$\hat{\vect q}_t$ and clamp the PD target to natural joint
limits.

\textbf{Input observation $\vect{o}_t$} decomposes into three parts, $\vect{o}_t=(\vect{o}^{\text{sim}}_t,\vect{o}^{\text{ref}}_t,\vect{a}_{t-1})$, where $\vect{o}^{\text{sim}}_t$ is the current simulator state, $\vect{o}^{\text{ref}}_t$ is a future reference goal, and $\vect{a}_{t-1}$ is the previously predicted residual correction term. To capture the hand-object interaction, the simulator state $\vect{o}^{\text{sim}}_t$ consists of hand poses $\vect{q}$, hand keypoints $\vect{K} \in\mathbb{R}^{2n_K\times 3}$, object states $\vect{P}_o$, contact indicators $\vect{c}\in\{0,1\}^{2n_B}$, and explicit displacement vectors $\vect{D}\in\mathbb{R}^{2n_K\times 3}$ from each hand keypoint to the nearest sample on the object surface, computed against a fixed canonical set $\{\vect{s}_k\}_{k=1}^{K}$ rigidly transformed into the world frame by the current object pose.

We construct the reference goal observation $\vect o^\text{ref}_t$ using the future mocap reference in a look-ahead horizon $\tau\in\mathcal{T}$ (sampled densely near $t$, sparsely out to $2$\,s). The goal $\vect{o}^{\text{ref}}_t$ stacks future mocap targets: reference object pose $\hat{\vect P}_{o, \tau}$, hand poses $\hat{\vect q}_{\tau}$, and keypoints $\hat{\vect K}_{\tau}$ and their corresponding residuals to the current state $\Delta \hat{\vect P}_{o, \tau}, \Delta \hat{\vect q}_\tau$, $\Delta \hat{\vect K}_\tau$ together with the reference binary contact map $\hat{\vect c}_\tau$ into the observation vector.
We use the 6D representation~\citep{zhou2020continuityrotationrepresentationsneural} for all rotations.

\noindent\textbf{Contact-aware tracking reward.}
We supervise the policy with a per-step reward that combines three tracking losses through exponential shaping: $r_t=\sum_{l\in\{L^{\text{samp}},L^{\text{cont}},L^{\text{key}}\}}\exp(-\lambda_l\,l)$
where the object surface-sample loss $L^{\text{samp}}$ tracks the rigid object pose, the contact loss $L^{\text{cont}}$ tracks the reference contact map, and the keypoint loss $L^{\text{key}}$ tracks the hand keypoint trajectory. We detail each loss next.

The \textit{object surface-sample loss} $L^{\text{samp}}=\tfrac{1}{K}\textstyle\sum_{k=1}^{K}\big\|\vect{P}_o\vect{s}_k-\hat{\vect{P}}_o\vect{s}_k\big\|_2$ captures object translation and rotation by tracking a fixed set of canonical object surface points $\{\vect{s}_k\}_{k=1}^K$ in the world frame.

The \textit{contact loss} $L^{\text{cont}}=\tfrac{1}{2n_B}\textstyle\sum_{j=1}^{2n_B}\max\!\big(\hat{\vect{c}}_j-\vect{c}_j,\,0\big)$ enforces reference contacts as a one-sided penalty on missed contacts. We deliberately do not penalize spurious contacts symmetrically: mocap contact maps are geometrically inferred and noisy, and a symmetric penalty would push the policy off physically plausible grasps.

For keypoint tracking, we combine two terms. The \textit{pose loss} $\ell_i^{\text{pose}}=\|\vect{P}_w^{-1}\vect{K}_i-\hat{\vect{P}}_w^{-1}\hat{\vect{K}}_i\|_2$ aims to minimize the distance between simulated and reference finger keypoints, expressed in their respective wrist frame, pushing the local pose of the hand to be consistent, independent of wrist deviations. The \textit{contact position loss} $\ell_i^\text{copo}=\|\vect{P}_o^{-1}\vect{K}_i-\hat{\vect{P}}_o^{-1}\hat{\vect{K}}_i\|_2$, expressed in object frames, drives the simulated fingers to meet the object at the same surface locations as the reference. These terms are combined into the final \emph{keypoint loss}  
$L^{\text{key}}=\tfrac{1}{2n_K}\textstyle\sum_{i=1}^{2n_K}\big(w_i\,\ell_i^{\text{copo}}+(1-w_i)\,\ell_i^{\text{pose}}\big)$, 
where the weighting $w_i$ shifts towards $\ell_i^\text{copo}$ as the keypoint approaches a reference contact event. Details on the computation of $w_i$ are given in Appendix~\ref{sec:impl_details}.

\noindent\textbf{Per-object experts and hard-mining.} Rather than training one expert per sequence, which is expensive, we train two generalist experts for each object where each expert handles either single or bimanual interactions respectively. 
We prune unrecoverable rollouts via early termination~\citep{peng2018deepmimic}: episodes end on hand self-penetration or on excessive deviation from the reference measured by $L^{\text{samp}}$, with the deviation threshold annealed as a curriculum. We further apply a hard-mining sampling strategy that biases the expert's sequence pool toward low recent tracking-completion rates. See Appendix~\ref{sec:impl_details} for details. With this training curriculum, our per-sequence experts outperform ManipTrans~\citep{li2025maniptrans}, and our per-object experts maintain a high success rate on the full dataset (\Cref{tab:eval-expert}).

\subsection{Stage II: Interaction Prior Distillation}\label{subsec:distillation_v2}
The experts trained in Stage I are privileged policies that can physically reproduce mocap interactions, but are not generative themselves, as they require full reference hand motions that are unavailable at generation time. 
The goal of Stage II is to compress the entire expert pool into a single goal-conditioned generative model whose \emph{learnable prior} and \emph{decoder} together act as an \emph{interaction-aware} prior over physically simulated hand-object motions.
This prior--decoder pair is then used in Stage III as the latent backbone for RL exploration. Unlike prior motion-generation works that model human kinematics~\citep{luo2024ominigrasp, tessler2024maskedmimic} without the object, our prior captures the coupled hand--object dynamics distilled from physically valid expert behavior.

\noindent\textbf{Object-trajectory-conditioned generative model.}
We instantiate the model as a conditional VAE~\citep{kingma2013auto, sohn15cvae}
with three components (see middle of \Cref{fig:method}) sharing a Gaussian latent space. The \emph{learnable prior}
$\mathcal{P}(\vect{z}_t\mid\tilde{\vect{o}}_t)$ is conditioned on the
\emph{partial observation} $\tilde{\vect{o}}_t$, which contains only the future
object-trajectory portion of the reference goal. The privileged
\emph{encoder} $\mathcal{E}(\vect{z}_t\mid\vect{o}_t)$ is conditioned on the
\emph{privileged observation} $\vect{o}_t$, which extends $\tilde{\vect{o}}_t$
with the reference hand trajectory features of the reference goal. 
The deterministic \emph{decoder} $\mathcal{D}(\vect o_t^{\text{sim}},\vect{z}_t)$
maps the current simulation state $\vect o_t^{\text{sim}}$ and a sampled latent to a per-step update on the simulated hand pose, which we apply as a residual on the current pose to form the PD target $\vect{Q}_t=\vect{q}_{t-1} + \mathcal{D}(\vect{o}^{\text{sim}}_t,\vect{z}_t)$. At training time, latents are drawn from the encoder,
$\vect{z}_t\sim\mathcal{E}(\cdot\mid\vect{o}_t)$; at test time the encoder is
discarded and we use the prior's mean $\mu_\mathcal P(\tilde{\vect o}_t)$ so generation never depends on reference hand poses.

\noindent\textbf{Training losses.}
We train our model with behavior-cloning loss $\mathcal{L}_{\text{BC}}$ and weighted Kullback-Leibler (KL) divergence loss $\mathcal{D}_{\text{KL}}$: $\mathcal{L}_{\text{distill}}=\mathcal{L}_{\text{BC}}+\beta\,\mathcal{D}_{\text{KL}}$.
Here $\mathcal{L}_{\text{BC}}$ matches the decoder output to the expert's PD target on every visited state. Rollouts follow DAgger~\citep{Ross2011A_Reduction_of}: we use the decoder to run simulation in the environment, while the experts are queried for a supervision action. At every time step, we match the student's PD target $\vect{Q}_t$ to the expert's PD target $\vect{Q}^*_t$ produced by the~\cref{subsec:tracking_v2} policy using $L_1$ loss: $\mathcal L_\text{BC} = \| \vect Q_t - \vect Q^*_t\|_1$. 
Matching at the PD-target level absorbs the reference-pose anchoring built into the expert's residual formulation, so the decoder learns natural per-step updates rather than reconstructing the anchoring offset.
The KL term 
$\mathcal{D}_{\text{KL}}=\mathrm{KL}\big(\mathcal{E}(\vect{z}_t\mid\vect{o}_t)\,\|\,\mathcal{P}(\vect{z}_t\mid\tilde{\vect{o}}_t)\big)$
aligns the partial-observation prior with the privileged-observation encoder, so that latents sampled from $\mathcal{P}$ at test time stay on the manifold the decoder was trained against.

\subsection{Stage III: Interaction Generation via Bounded Latent Exploration}\label{subsec:generative_tasks_v2}
Stage~II distillation yields a compact, interaction-aware latent space whose samples decode to physically plausible actions. To generalize beyond the limited training objects and interactions to novel objects, trajectories, and tasks, we explore this space with RL. The learned interaction prior improves sample efficiency and suppresses unnatural motion, providing a general approach to different manipulation tasks (\cref{sec:experiments}).

\noindent\textbf{Bounded latent exploration setup.}
Independent of the concrete task, we train an exploration policy $\pi_\text{explore}$ via PPO that consumes the partial observation $\tilde{\vect{o}}_t$ and the current simulator state $\vect{o}^{\text{sim}}_t$, and outputs a residual latent $\Delta\vect{z}_t$ on top of the frozen prior's mean. The action is then produced by the frozen decoder $\vect{a}_t=\mathcal{D}\big( \,\vect{o}^{\text{sim}}_t, \,\mu_\mathcal{P}(\tilde{\vect{o}}_t)+\Delta\vect{z}_t\big)$. To keep the latent close to the original space, we bound the magnitude of $\Delta \vect z_t$ by a constant $B$ (similar to \citealp{yu2026zprl, zhang2026learningathletichumanoidtennis}) via a smooth squash $\Delta \vect z \leftarrow B \tanh\big(\frac{\|\Delta \vect z\|}{B})\frac{\Delta \vect z}{\|\Delta \vect z\|}$. We show in \cref{sec:ablations} that this bound constrains the exploration and maintains synthesis diversity.

\section{Experiments}
\label{sec:experiments}

We compare our Stage~I experts with a state-of-the-art mocap tracker and our interaction-prior generator with baselines on tabletop manipulation, object catching, and bimanual handover, consistently improving success rates. Videos and interactive qualitative results are on our \href{https://virtualhumans.mpi-inf.mpg.de/PiHand/}{project page}.

\noindent\textbf{Datasets and Implementation Details.}
We use GRAB~\citep{GRAB:2020} as our dataset for training the interaction prior, but also rely on DexYCB~\citep{dexycb:cvpr2021} as a secondary source to show the cross-dataset generalization capabilities of our GRAB-trained interaction prior. 
All experiments are done within IsaacGym~\citep{makoviychuk2021isaacgymhighperformance} with MANO-consistent robotic hands. Further details on data split, preprocessing and simulator can be found in Appendix~\ref{app: exp}.

\textbf{Evaluation protocol and metrics.}
Unless otherwise stated, we evaluate each method using 100 simulated rollouts per sequence to account for simulator stochasticity. A rollout is terminated when the object deviates from the reference trajectory beyond a task-dependent distance threshold. To measure temporal robustness, we report \emph{tracking completion (TC in $\%$)}, defined as the fraction of the reference trajectory completed before termination, and \emph{success rate (SR in $\%$)}, defined as the fraction of rollouts that complete the entire sequence. For spatial precision, we report average \emph{object translation error ($E_t$ in cm)}, the \emph{object rotation error ($E_r$ in $^\circ$)}, and, for the Stage I imitation task where reference motions are available, the \emph{mean per joint position error (MPJPE in cm) } of the hands.

\begin{figure}[t]
    \centering
    \includegraphics[width=\linewidth]{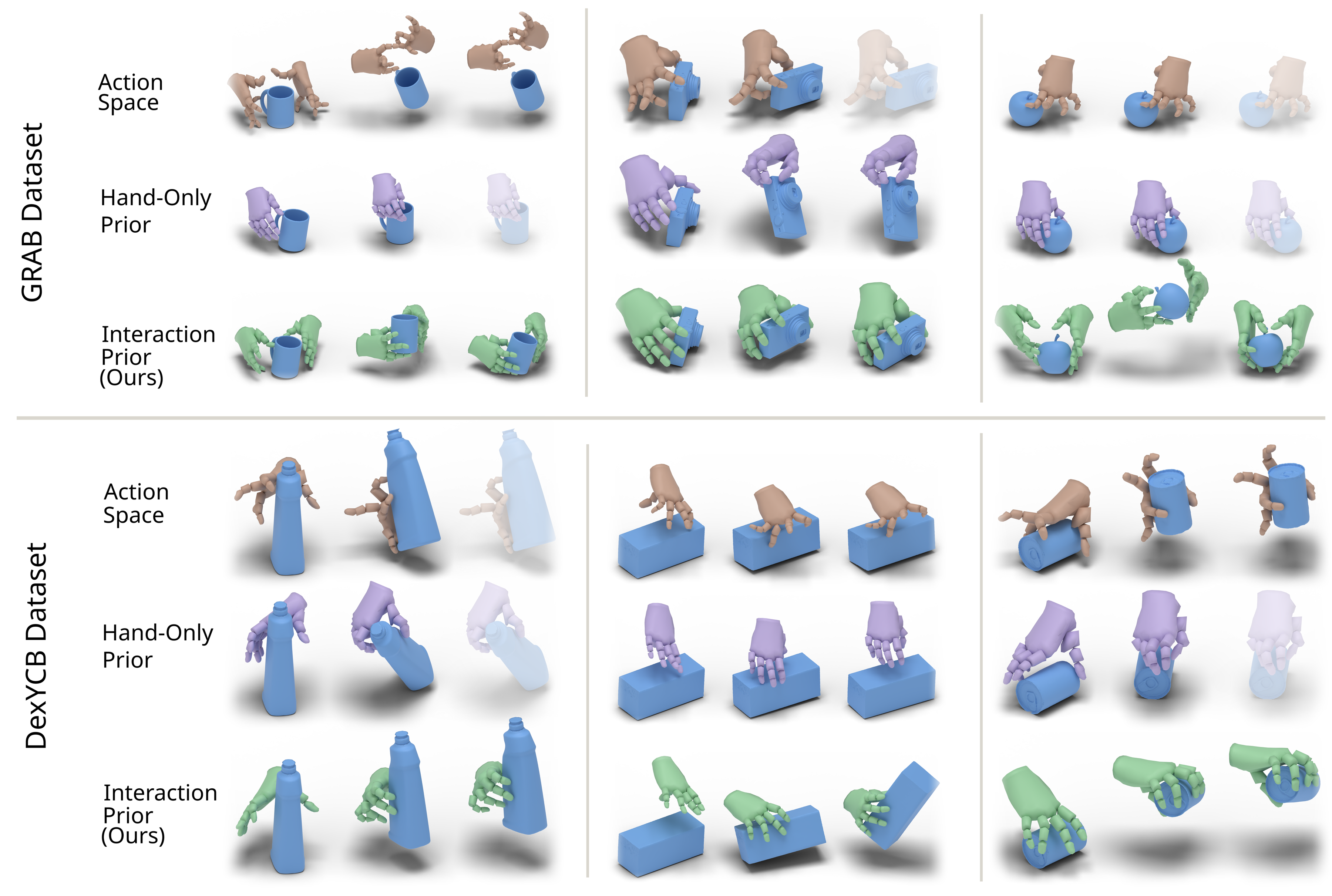}
    \caption{\textbf{Tabletop manipulation synthesis comparison} on unseen objects from GRAB (top row) and DexYCB (bottom row) evaluation set. Grayed out frames indicate early termination due to fatally wrong grasps. Action space exploration generates unnatural grasping and hand-only prior often fails to grasp with object reference while ours achieves both realism and high success rate.}
    \label{fig:compare-synthesis}
\end{figure}

\subsection{Hand--Object Interaction Imitation}
\label{sec:exp_tracking}

We compare our Stage~I experts with ManipTrans~\citep{li2025maniptrans}, a state-of-the-art per-sequence imitation method.
Since ManipTrans does not support bimanual manipulation of single objects, we compare with it on GRAB50 which is a random subset of 50 single-handed GRAB~\citep{GRAB:2020} sequences. 
As \Cref{tab:eval-expert} shows, our experts trained per-sequence outperform ManipTrans trained on an identical setup on all metrics. We also train two experts per object, one for single-handed and one for bimanual interactions, on the full GRAB dataset. While the success rate slightly degrades, it is much more scalable than training one expert per sequence as done in ManipTrans and our overall success rate is still better than per-sequence ManipTrans experts. This shows that our method is a scalable data engine for learning physical interaction priors from human demonstrations. 

\begin{table*}[h]
    \centering
    \caption{\textbf{Physical imitation of mocap reference motions} on GRAB50 and the full GRAB dataset. Our method significantly improves the tracking success rate than ManipTrans~\citep{li2025maniptrans}.}
    \label{tab:eval-expert}
    \small
    \setlength{\tabcolsep}{5pt}
    \renewcommand{\arraystretch}{1.1}
    \resizebox{\linewidth}{!}{%
    \begin{tabular}{l l l c c c c c}
    \toprule
    Method & Data & Granularity & SR $\uparrow$ & TC $\uparrow$  & $E_t$ (cm)$\downarrow$ & $E_r$ ($^\circ$)$\downarrow$  & MPJPE (cm) $\downarrow$\\
    \midrule
    ManipTrans
        & GRAB50 & Per-sequence & 69.6 & 77.6 & 0.43 & 7.3 & 1.97 \\
    \midrule
    \multirow{2}{*}{Ours}
        & GRAB50 & Per-sequence & \textbf{93.3} & \textbf{95.8} & \textbf{0.28} & \textbf{4.9} & \textbf{1.01} \\
        & GRAB Full & Per-object & 84.8 & 92.2 & 1.04 & 15.4 & 2.08 \\
    \bottomrule
    \end{tabular}
    }
\end{table*}

\subsection{Tabletop Manipulation}
\label{subsec:tabletop}

Our main application of the interaction prior is a task we call \emph{Tabletop Manipulation}: given only a target object trajectory, a single policy must synthesize a physically plausible interaction for objects and trajectories it has never seen.
To this end, we train the Stage~III residual policy $\pi_\text{explore}$ over a broad distribution of initial states and target trajectories.

\noindent\textbf{Setup.}
Each training episode initializes the hands and the object from two \emph{different} training sequences, so the policy starts from hand--object configurations that neither the experts nor the prior have encountered.
The target trajectory is, with equal probability, an original training trajectory or a procedurally generated lift-and-move trajectory.
The policy only ever interacts with training objects, whereas at evaluation both objects and trajectories are held out, making the policy zero-shot with respect to object geometry and motion.
We train one Stage~III policy per dataset, on GRAB and on DexYCB, each on the respective training objects and on top of the same GRAB-trained prior.
Since generation is considerably harder than imitation, we relax the termination threshold for SR and TC to 10\,cm. However, our advantage persists and even grows under stricter thresholds (Appendix~\ref{app:threshold}).

\noindent\textbf{Baselines.}
We implement two alternatives within our framework, so that simulator, embodiment and physics are identical across methods (see Appendix~\ref{sec:appendix_baselines}).
\textbf{Action Space Exploration} lets $\pi_\text{explore}$ act directly in the 102-dimensional joint space without use of a learned prior.
\textbf{Hand-Only Prior} is similar to Omnigrasp~\citep{luo2024ominigrasp} and PULSE~\citep{Luo2024PULSE} which lets $\pi_\text{explore}$ act in a latent space distilled from experts that track hand motion without the object.

\noindent\textbf{Results.}
To accelerate baseline learning, we additionally give both baselines a pre-grasp reward~\citep{dasari2023learning,luo2024ominigrasp} and a difficulty curriculum (\Cref{app:additional_rewards_curriculum}).
Our method needs neither and is trained on the full task distribution from the start.
As Table~\ref{tab:generative_evaluation} shows, it nonetheless outperforms both baselines on unseen objects, improving the success rate over the strongest baseline by 14.0 points on GRAB and 16.9 points on DexYCB while also reducing object pose errors.
It also reaches the peak baseline performance with only 1/10 of the training time (in 900 epochs instead of 8900 epochs), demonstrating much higher training efficiency.
Because the DexYCB policy builds on the prior trained exclusively on GRAB, these results further show that the interaction prior generalizes across datasets. Qualitative results from both datasets are depicted in \Cref{fig:compare-synthesis}.

\begin{table*}[htb]
    \centering
    \caption{\textbf{Tabletop manipulation synthesis} on GRAB and DexYCB with unseen objects and trajectories. The interaction prior is trained only on GRAB dataset.}
    \label{tab:generative_evaluation}
    \small
    \setlength{\tabcolsep}{4pt}
    \renewcommand{\arraystretch}{1.1}
    \begin{tabular}{l cccc @{\hspace{3ex}} cccc}
        \toprule
        & \multicolumn{4}{c}{GRAB~\citep{GRAB:2020}} & \multicolumn{4}{c}{DexYCB~\citep{dexycb:cvpr2021}} \\
        \cmidrule(r{3ex}){2-5} \cmidrule(l){6-9}
        Method & SR $\uparrow$ & TC $\uparrow$  & $E_t$ (cm)$\downarrow$ & $E_r$ ($^\circ$)$\downarrow$ & SR $\uparrow$ & TC $\uparrow$  & $E_t$ (cm)$\downarrow$ & $E_r$ ($^\circ$)$\downarrow$ \\
        \midrule
        Action Space Expl. & 59.1 & 73.6 & 1.79 & 21.5 & 60.8 & 81.3 & 1.63 & 14.2 \\
        Hand-Only Prior & 60.0 & 73.4 & 1.52 & 21.2 & 59.5 & 79.4 & 1.59 & 16.1 \\
        Ours & \textbf{74.0} & \textbf{82.9} & \textbf{1.36} & \textbf{20.0} & \textbf{77.7} & \textbf{87.4} & \textbf{1.52} & \textbf{13.8} \\
        \bottomrule
    \end{tabular}
\end{table*}
\begin{figure}
  \centering
  \captionsetup[subfigure]{justification=centering, skip=1pt}
  \captionsetup[sub]{subrefformat=parens}
  \begin{subfigure}[b]{0.32\linewidth}
    \includegraphics[width=\linewidth]{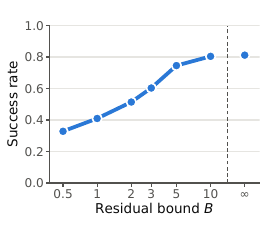}
    \caption{Success rate}\label{fig:bound_sr}
  \end{subfigure}\hfill
  \begin{subfigure}[b]{0.32\linewidth}
    \includegraphics[width=\linewidth]{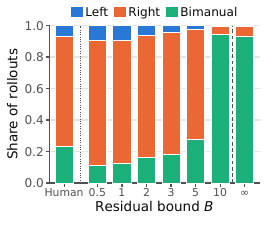}
    \caption{Hand usage}\label{fig:bound_hands}
  \end{subfigure}\hfill
  \begin{subfigure}[b]{0.32\linewidth}
    \includegraphics[width=\linewidth]{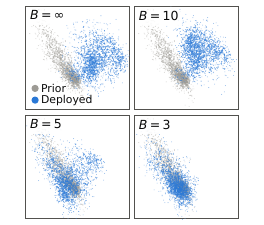}
    \caption{Latent space (PCA)}\label{fig:bound_pca}
  \end{subfigure}
  \caption{\textbf{Effect of the latent residual bound $B$}. Exploration with less constraint (larger bound value B) leads to higher success rate \subref{fig:bound_sr} but collapsed grasp diversity \subref{fig:bound_hands} as the deployed latents (blue) deviate too much from the learned latent prior (grey) in \subref{fig:bound_pca}. 
  }
  \label{fig:bound_ablation}
\end{figure}

\subsection{Generalization to New Tasks}
\label{sec:downstream}
\newsavebox{\newtasktab}
\sbox{\newtasktab}{\footnotesize	\setlength{\tabcolsep}{4pt}\renewcommand{\arraystretch}{1.1}%
  \begin{tabular}{lcc}
    \toprule
    \footnotesize	
    Method & Catching & Handover \\
    \midrule
    Action Space Expl. & 43.0 & 11.9 \\
    Hand-Only Prior    & 1.0  & 5.0  \\
    Ours (right hand)  & 78.0 & --   \\
    Ours               & \textbf{87.0} & \textbf{51.9} \\
    \bottomrule
  \end{tabular}}
\begin{wraptable}{R}{\wd\newtasktab}
  \centering
  \vspace{-1em}
  \captionsetup{justification=raggedright,singlelinecheck=false,skip=3pt}
  \caption{Success rate (\%$\uparrow$) on held-out GRAB objects for new tasks. 
  }
  \label{tab:catch_evaluation}\label{tab:handover_evaluation}
  \usebox{\newtasktab}
  \vspace{-1em}
\end{wraptable}

The tabletop manipulation policy above is trained on similar interactions to those the prior was trained on. We now ask whether the interaction prior transfers to tasks whose dynamics are absent or very rare in the training data. We consider two: \emph{catching} a thrown object, and a bimanual \emph{handover} in which the object must be passed from the left hand to the right. Both are formulated as object-trajectory-conditioned manipulation generation, so the exploration policy $\pi_\text{explore}$, the object split, and the baselines are unchanged from \Cref{subsec:tabletop}. The object-tracking reward is complemented by simple task-specific terms. We show success rate in \cref{tab:catch_evaluation} and qualitative examples in \cref{fig:downstream}. 
Task definitions, implementation details and full result tables are given in Appendices~\ref{app:downstream_common}--\ref{app:downstream_handover}. 

\noindent\textbf{Catching.}
An object with random orientation and spin is launched on a ballistic arc towards an intercept point. The hands start open, palms facing, at a random offset (\cref{app:downstream_catch}). We define both the bimanual- and right-hand catching task, where we encourage different hands to catch the thrown object. \cref{tab:catch_evaluation} shows that our interaction prior is highly beneficial, achieving $2.0\times$ the success rate of the strongest baseline. 

\noindent\textbf{Handover.}
This task is designed to force a handover of the object from the left to the right hand. The object is initialized in the left hand, and the reference trajectory moves it in a straight line to the right side of the workspace. Each hand is allowed to stay only half of the workspace, ensuring that object movement is only possible by a proper change of hands. \cref{tab:handover_evaluation} shows that our interaction prior is critical for this challenging task with $4.3\times$ the success rate of the strongest baseline. 

\begin{figure}[h]
    \centering
    \includegraphics[width=\linewidth]{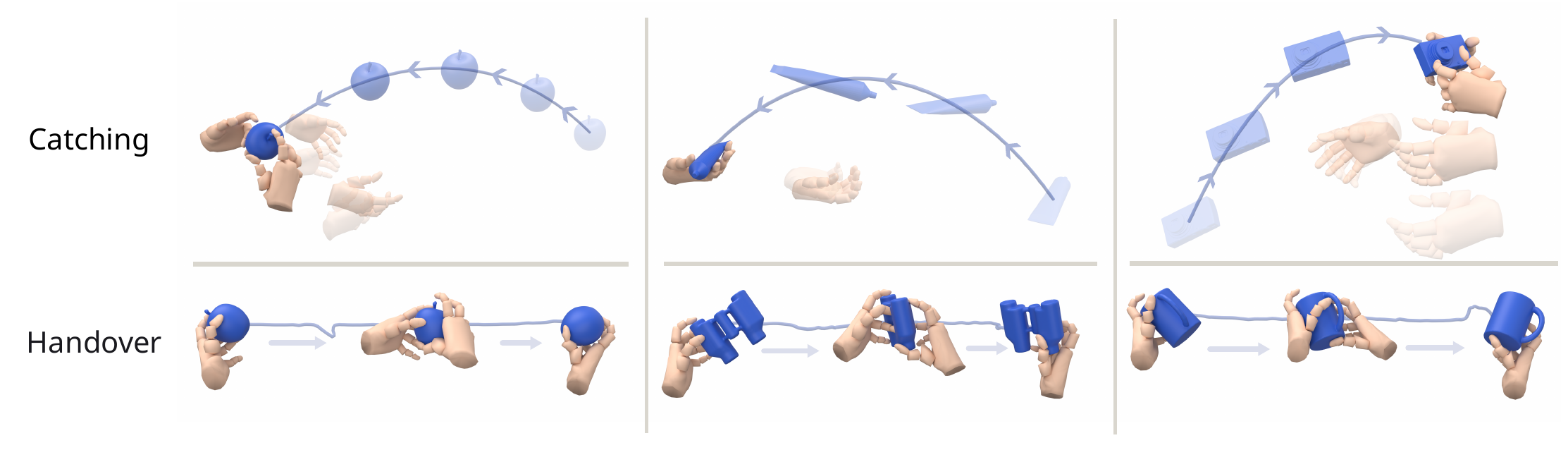}
    \caption{\textbf{Generalization of the interaction prior} to object catching (top) and handover (bottom). Our interaction prior transfers to other manipulation tasks rarely seen in GRAB.
    }
    \label{fig:downstream}
\end{figure}

\subsection{Ablations}
\label{sec:ablations}

\newsavebox{\abltab}
\sbox{\abltab}{\footnotesize\setlength{\tabcolsep}{4pt}\renewcommand{\arraystretch}{1.1}%
  \begin{tabular}{lc}
    \toprule
    Method & SR $\uparrow$ \\
    \midrule
    Direct prior rollout    & 30.8 \\
    Action-space residual   & 49.6 \\
    w/o random traj.        & 67.9 \\
    \midrule
    Ours                    & \textbf{74.0} \\
    \bottomrule
  \end{tabular}}

\noindent
\begin{minipage}[t]{\dimexpr\linewidth-\wd\abltab-1.5em\relax}
  We run all ablations on Tabletop Manipulation on GRAB.
  \par\smallskip
  \textbf{Stage~III components.}
  \Cref{tab:ablations} ablates three components of Stage~III, which we define formally and show full table in Appendix~\ref{app:ablations}.
  Executing the frozen prior without an exploration policy reaches only 30.8\% success, so Stage~III more than doubles the performance of the distilled prior.
  Adding the residual to the decoded action instead of the latent lowers the success to 49.6\%, showing that exploring within the prior's latent space is substantially more effective than correcting its actions. Random trajectory augmentation also boosts the success rate by about six points.
\end{minipage}\hfill
\begin{minipage}[t]{\wd\abltab}
  \captionsetup{type=table,justification=raggedright,singlelinecheck=false,skip=3pt}
  \caption{Stage~III ablations, success rate on held-out GRAB objects. 
  }
  \label{tab:ablations}
  \usebox{\abltab}
\end{minipage}

\noindent\textbf{Residual bound.} We introduce a bound in the residual latent to constrain the RL exploration in stage III. We analyze how this affects success rate and grasp diversity in \cref{fig:bound_ablation}. Success rises as the bound is relaxed and up to $B{=}5$ the deployed latents stay within the prior's distribution and hand usage follows the human reference. Looser bounds improve success only marginally, but the latents drift away from the prior and the policy collapses almost entirely to bimanual grasps, so we use $B{=}5$. 

\noindent\textbf{Robustness.}
Appendix~\ref{app:robustness} evaluates robustness to changes in friction, object mass, and action noise.
Our policy retains the highest success rate under all perturbations, and degrades far less than the baselines when physical object properties deviate.

\section{Conclusion}

We propose \method{}, a physics-based framework for synthesizing dexterous hand--object interactions from target object trajectories.
Our key idea is to learn an explicit interaction prior that jointly models hand motion, object geometry, and contact dynamics, rather than relying on hand-only priors or reward-driven contact discovery.
\method{} trains imitation experts on mocap data, distills them into a conditional VAE prior, and explores its latent space to solve novel tasks.
Our experts improve SR over ManipTrans from 69.6\% to 93.3\% and maintain 84.8\% SR on the full GRAB set with per-object experts.
On unseen GRAB and DexYCB objects, our tabletop manipulation policies achieve 14.0 and 16.9 points higher SR than the strongest baselines while learning roughly ten times faster. The same prior transfers to catching and handover, tasks rare or absent in its training data, achieving $2.0\times$ and $4.3\times$ the strongest-baseline SR, respectively.
We will release the processed data, expert policies, and generation model.

\noindent\textbf{Limitations.} While \method{} clearly improves over prior baselines, several limitations remain. First, our tracking experts do not yet reach 100\% success rate on the full GRAB dataset. Failures concentrate on extremely dynamic motions and sequences with heavy mocap noise, which we expect could be addressed with stricter termination criteria, curriculum learning, or higher-quality reference data. Second, we focus on rigid objects and do not model articulated objects such as those in ARCTIC~\citep{fan2023arctic}. Extending the interaction prior to articulated geometry is an important next step. Third, our framework controls only the hands and does not consider full-body motion. Combining a full-body locomotion controller with our hand manipulation prior is a promising future direction. Finally, all experiments are conducted in simulation with a single MANO-based hand model. Transferring \method{} to real robotic hands would be an exciting direction, but it requires retargeting the prior to a different embodiment and bridging the gap between simulated and real sensing and actuation.

\subsubsection*{Acknowledgments}
The project was made possible by funding from the Carl Zeiss Foundation. This work was supported by the German Federal Ministry of Education and Research (BMBF): Tübingen AI Center, FKZ: 01IS18039A. This work is funded by the Deutsche Forschungsgemeinschaft (DFG, German Research Foundation) - 409792180 (Emmy Noether Programme, project: Real Virtual Humans). Gerard Pons-Moll is a member of the Machine Learning Cluster of Excellence, EXC number 2064/1 -- Project number 390727645. The authors thank the International Max Planck Research School for Intelligent Systems (IMPRS-IS) for supporting Enes Duran.

\bibliography{main}
\bibliographystyle{iclr2027_conference}

\vfill

\pagebreak
\appendix
\section{Implementation Details}
\label{sec:impl_details}

This section provides the network architectures, hyperparameters and termination criteria needed to reproduce the three stages of our pipeline, starting with details shared by all stages.

\subsection{Shared Details}
\label{app:shared_details}

\noindent\textbf{Observation horizon.}
In all stages, the observation contains future frames of the object reference trajectory and, where applicable, future mocap hand poses. Stages~I and~III sample these frames at the offsets $\mathcal T = \{ 1, 2, 3, 4, 6, 8, 12, 16, 24, 32, 48, 64, 96, 128\}$. To avoid overfitting of the latent prior, Stage~II uses the reduced set $\mathcal T = \{ 1, 2, 4, 8, 16, 32, 64\}$.

\noindent\textbf{Self-collisions.}
In IsaacGym we disable all collisions between the rigid bodies within one hand model. The simulator cannot identify rigid-body contact pairs and only provides the total contact force acting on each rigid body. Since we infer the binary contact maps $\vect{c}_t$ from these contact forces, we must ensure that they stem from the hand colliding with the scene and not with itself. To prevent the resulting unfiltered self-penetrations of the hand, we check for self-penetration by approximating the fingers as cylinders of 1\,cm diameter, and use these checks as a termination criterion in all stages of the pipeline.

\subsection{Stage I: Expert Training}
\label{app: expert training}

\noindent\textbf{Keypoint weighting.}
The keypoint loss $L^{\text{key}}$ tracks all $2n_K$ keypoints in a frame that interpolates between wrist-local (non-contact) and object-centric (in-contact) via the time-to-contact weight $w_i =\exp(-\tau_i/\tau_0)$, where $\tau_i$ is the number of frames until keypoint $i$ next contacts the object in the reference mocap trajectory, and $\tau_0=30$.

\noindent\textbf{Hard-mining sampling.}
We sample reference sequence $i$ from the expert's pool with a probability that interpolates between uniform and failure-proportional sampling:
\begin{equation}
P(i)=(1-\alpha)\tfrac{1}{N}+\alpha\,\tfrac{1-\text{TC}_i+\epsilon}{\sum_{j=1}^{N}(1-\text{TC}_j+\epsilon)},
\label{eq:hard_mining}
\end{equation}
with pool size $N$, recent tracking-completion rate $\text{TC}_i\in[0,1]$, hard-mining strength $\alpha$, and numerical-stability constant $\epsilon=10^{-6}$. This ensures that harder sequences are sampled more frequently than those that are already tracked well.

\noindent\textbf{Networks and optimization.}
The actor is a 3-layer MLP with hidden layer sizes (1024, 1024, 512), and the critic is a 2-layer MLP with layer sizes (1024, 512). Both networks use GELU activations. All PPO hyperparameters are summarized in \Cref{tab:expert_hyperparameters}.

\begin{table}[h]
    \centering
    \caption{Hyperparameters for expert policy training. Reward weights and environment thresholds are shared across all expert variants. We scale the minibatch size with the dataset diversity of the respective training task.}
    \label{tab:expert_hyperparameters}
    \small
    \renewcommand{\arraystretch}{1.1}
    \begin{tabular}{lc}
        \toprule
        \textbf{Parameter} & \textbf{Value} \\
        \midrule
        \multicolumn{2}{c}{\textit{Reward Weights \& Environment}} \\
        \midrule
        Keypoint tracking weight ($\lambda_\text{key}$) & $10$ \\
        Object surface-sample weight ($\lambda_\text{samp}$) & $10$ \\
        Contact tracking weight ($\lambda_\text{cont}$) & $2.5$ \\
        Termination threshold: Object position deviation & 5\,cm $\rightarrow$ 2\,cm (annealed) \\
        Termination threshold: Object sample deviation & 5\,cm $\rightarrow$ 2\,cm (annealed) \\
        Action standard deviation & Learnable, initialized at $\exp(-1)$ \\
        \midrule
        \multicolumn{2}{c}{\textit{Optimization \& PPO Settings}} \\
        \midrule
        Optimizer & Adam \\
        Learning rate & $2 \times 10^{-5}$ \\
        Number of environments & 4096 \\
        Rollout length (steps per env) & 32 \\
        Discount factor ($\gamma$) & 0.99 \\
        GAE parameter ($\tau$) & 0.95 \\
        Clip range ($e_\text{clip}$) & 0.2 \\
        PPO mini-epochs & 6 \\
        Minibatch size (ManipTrans comparison) & 1024 \\
        Minibatch size (Per-object experts) & 16384 \\
        \bottomrule
    \end{tabular}
\end{table}

\subsection{Stage II: Distillation}
\label{app:distillation}

Encoder, prior and decoder are MLPs with layer sizes (2048, 2048, 1024, 1024, 512), GELU activations and layer normalization, and share a latent space of dimension 64. All spatial observations, such as the object pose, are passed to these networks expressed in the wrist frames of \emph{both} hands, which makes them invariant to global scene transformations. Following~\cite{luo2024ominigrasp}, we parameterize the encoder mean as a residual on the prior mean, $\mu_\mathcal{E}=\mu_\mathcal{P}(\tilde{\vect{o}}_t)+\Delta\mu(\vect{o}_t)$, where the encoder network outputs the residual $\Delta \mu(\vect o_t)$. We linearly anneal the KL loss weight from 0 to $10^{-4}$ and additionally apply episodic latent noise~\citep{tessler2024maskedmimic} to prevent posterior collapse. All networks are trained with Adam and a learning rate of $10^{-4}$. To avoid overfitting of the prior to the training sequences, we remove all velocities from the simulation state observation $\vect{o}_t^{\text{sim}}$, represent future reference poses only relative to the current state, and use a shorter look-ahead (see Appendix \ref{app:shared_details}). This improved the direct prior rollout success rate on the held-out evaluation set from 23.7\% to 30.8\%, when compared with a prior trained on full observations.

The simulation setup is identical to Stage~I, and we employ the same hard-mining strategy. We terminate environments early when the object position deviates by more than 5\,cm from the reference, since the experts never encountered larger deviations during their own training and their supervision would be unreliable in such states.

\subsection{Stage III: Latent Exploration}
\label{app:latent_exploration}

The latent exploration policy is an MLP with hidden layer sizes (2048, 1024, 1024, 512) and ReLU activations, and we bound its latent residual with $B=5$. For Tabletop Manipulation, the only reward term is $\exp(-10\,L^\text{samp})$, based on the object surface-sample loss from Stage~I; the rewards for the downstream tasks are described in \Cref{app:downstream_common}. Episodes terminate early when the object deviates by more than 10\,cm from its reference position, which matches the evaluation threshold. The action standard deviation is initialized to 1, and all remaining optimization and PPO settings are the same as in Stage~I.

\section{Data and Simulation}
\label{app: exp}

\subsection{Datasets and Preprocessing}
\label{app:datasets}

\noindent\textbf{GRAB.}
We use the GRAB~\citep{GRAB:2020} dataset as the primary source of interaction demonstrations. GRAB contains high-quality whole-body human grasping motions with detailed hand poses and contact annotations, collected from 10 subjects interacting with 51 everyday objects. We use 50 objects in our experiments, excluding \texttt{doorknob} due to physically implausible floating trajectories in the source data. Following prior work~\citep{Taheri:CVPR:2022:GOAL}, we partition the dataset into a training and a cross-object evaluation split. Raw motion capture sequences often contain extensive segments without actual object manipulation, such as the initial arm approach or the post-release retraction. To focus our policies on the core interaction dynamics, we temporally trim the sequences to segments that begin approximately 40 frames before the initial contact and end 40 frames after the contact is broken. Finally, to prevent lateral bias and increase data diversity, we apply a left-right mirroring augmentation to all sequences. After preprocessing, the final dataset contains 4,854 training sequences with an average length of 317.4 frames at 60\,Hz.

\noindent\textbf{DexYCB.}
We use DexYCB~\citep{dexycb:cvpr2021} to evaluate generalization beyond GRAB. DexYCB contains sequences of 10 subjects grasping 20 objects from the YCB object set with a single hand, annotated with MANO hand poses and 6D object poses. In each sequence, a subject picks up a single target object from a table among several distractor objects and lifts it. We use the 250 right-handed sequences of subjects 1--5, covering all 20 objects. Following the unseen-object setting (S3) of DexYCB, we train on only 15 objects (186 sequences) and hold out 5 objects for evaluation (\texttt{005\_tomato\_soup\_can}, \texttt{009\_gelatin\_box}, \texttt{021\_bleach\_cleanser}, \texttt{036\_wood\_block}, \texttt{061\_foam\_brick}; 64 sequences). As for GRAB, we temporally trim the sequences to the interaction. Because all sequences are right-handed, we embed them into our bimanual representation with the left hand idle and do not apply mirroring augmentation. After preprocessing, the sequences have an average length of 196.4 frames at 60\,Hz.

\subsection{Simulation Setup and MANO Integration}
\label{app:simulation}

To bridge MANO-based human demonstrations and physical simulation, we construct our simulated robotic hand directly from the MANO model~\citep{MANO:SIGGRAPHASIA:2017}. The simulated hand follows the kinematic joint structure of the mean-shape MANO template in a flat-hand pose, with rigid-body collision geometries obtained by segmenting the corresponding MANO mesh. This design preserves a direct correspondence between the simulated and the visual hand representation, so that simulated trajectories can be mapped back to MANO without error-prone kinematic retargeting. While the GRAB dataset contains motions from 10 different human subjects with varying hand shapes, we use this single template hand across all environments. Rather than retargeting each demonstration to the template, we let the Stage~I tracking policy learn to compensate for the remaining morphology mismatch during RL training.

For all simulated objects, we set the volumetric density to 500\,kg/m\textsuperscript{3} and the contact friction coefficient to 0.7. Prior methods often use elevated friction values that artificially simplify grasping, such as 1.0 in Omnigrasp~\citep{luo2024ominigrasp} and 1.5 in MaskedManipulator~\citep{Tessler2025MaskedManipulator}. Our lower friction prevents unrealistic surface adhesion and encourages the policy to learn physically plausible force-closure grasps. Simulation and policy execution are synchronized at 60\,Hz. Since IsaacGym assumes convex meshes for efficient collision detection, we decompose all object meshes into convex parts using~\cite{wei2022approximate}.

\subsection{Training Times}
\label{app:training_times}

All models are trained on a single Nvidia A100. Each Stage~I expert is trained for up to 6\,h, however, on average they converge after only 1 or 2\,h. Distillation into the latent interaction prior takes 24\,h. Training the Tabletop Manipulation policy takes up to 36\,h, while the catching and handover policies are trained for 20\,h.

\section{Baselines}
\label{sec:appendix_baselines}

Several existing methods qualify as baselines for generative hand--object interaction, most notably full action space exploration frameworks such as GraspXL~\citep{Zhang2025GraspXL}, and human-motion latent exploration methods such as Omnigrasp~\citep{luo2024ominigrasp} and the work ~\cite{braun2023physicallyplausiblefullbodyhandobject}.

However, conducting fair, direct comparisons using official codebases is challenging for several reasons. First, some methods lack publicly available or well-documented code, making reproduction outside of their original environment practically impossible without major engineering efforts. Second, existing methods leverage vastly different training datasets. Crucially, differences in the underlying physics simulation---such as the choice of engine, friction coefficients, object mass densities, and the specific embodiment---fundamentally alter the difficulty of the task. For instance, full-body methods~\citep{braun2023physicallyplausiblefullbodyhandobject, luo2024ominigrasp} approximate an SMPL-X~\citep{SMPL-X:2019} body using coarse geometric primitives (capsules and cuboids), leading to radically different contact dynamics than high-fidelity, MANO-based hand interactions.

To isolate the impact of the interaction-first prior and ensure a strictly fair comparison, we implement the paradigms of these baselines directly within our framework. By standardizing the simulator, physics parameters, and embodiment across all methods, any observed difference in performance can be attributed to the approaches themselves rather than to environmental discrepancies. In the following we describe the concrete details of our baseline implementations.

\subsection{Action Space Exploration}
\label{app:baseline_action_space}

The action space exploration baseline represents the unconstrained RL approach. Here, the policy $\pi_\text{explore}$ directly outputs continuous actions in the 102-dimensional PD-target residual space. No generative prior is learned from human demonstrations; instead, the policy must discover grasping mechanics relying purely on the task-specific reward signal. This mirrors the exploration paradigm of frameworks such as GraspXL~\citep{Zhang2025GraspXL}.

\subsection{Hand-Only Prior}
\label{app:baseline_hand_only}

Methods like Omnigrasp~\citep{luo2024ominigrasp} improve upon raw action space exploration by first compressing human imitation skills into a proprioceptive, motion-only latent space, which is subsequently explored to solve object interactions. We reimplement this paradigm as our Hand-Only Prior baseline. To distinguish the observations in this motion-only setting from our interaction-aware setting ($\vect{o}_t, \tilde{\vect{o}}_t$), we denote them as $\vect{x}_t$.

\subsubsection{Hand-Only Imitation Experts}
We first train tracking experts exclusively on the hand kinematics of the GRAB dataset, entirely ignoring the object. The main differences to our interaction expert training are outlined below.

\noindent\textbf{Observation space.}
The policy observation is $\vect{x}_t = (\vect{x}^{\mathrm{sim}}_t,\,\vect{x}^{\mathrm{ref}}_t,\,\vect{a}_{t-1})$. Because the environment contains neither a table nor a target object, the state dimensions are significantly reduced. The simulation state consists only of the hand root poses, joint angles, and keypoints and their velocities, $\vect{x}^\text{sim}_t = (\vect{q}, \dot{\vect{q}}, \vect{K}, \dot{\vect{K}})$, and the reference observation omits the object trajectory, $\vect{x}^{\mathrm{ref}}_t = \big\{ \hat{\vect{q}}_{\tau},\,\Delta\hat{\vect{q}}_{\tau},\; \hat{\vect{K}}_{\tau},\,\hat{\vect{K}}_{\tau}-\vect{K} \big\}_{\tau\in\mathcal{T}}$.

\noindent\textbf{Reward formulation.}
The policy is rewarded solely for minimizing the distance to the reference keypoints, using the single reward term $\exp \big( - \tfrac{\lambda}{2n_K}\sum_{j=1}^{2n_K} \| \vect{K}_j - \hat{\vect{K}}_j \|_2 \big)$ with $\lambda = 100$.

\noindent\textbf{Training details.}
Without contact dynamics to model, this imitation task is much simpler, so we do not require per-object experts. Instead, we partition the dataset into 4 subsets and train one expert per subset. Episodes terminate early upon hand self-penetration or if the average keypoint error exceeds 2.5\,cm, and we use the same hard-mining strategy as for our interaction experts. As \Cref{tab:motion_experts} shows, the hand-only experts achieve high success rates and a very low MPJPE even under this strict threshold, so the baseline prior is distilled from well-trained experts.

\begin{table}[h]
    \centering
    \small
    \caption{Hand-only imitation experts used to build the Hand-Only Prior baseline. SR and TC in \% under a 2.5\,cm keypoint error threshold, MPJPE in cm.}
    \label{tab:motion_experts}
    \setlength{\tabcolsep}{5pt}
    \renewcommand{\arraystretch}{1.1}
    \begin{tabular}{l l c c c}
        \toprule
         & Dataset & SR $\uparrow$ & TC $\uparrow$ & MPJPE $\downarrow$ \\
        \midrule
        Hand-only imitation experts & GRAB + mirroring & 79.5 & 84.5 & 0.53 \\
        \bottomrule
    \end{tabular}
\end{table}

\subsubsection{Distillation into a Hand-Only Latent Space}
We keep the conditional VAE architecture with a learnable prior, but remove all object-related information from its inputs. The partial observation is simply the simulated hand state, $\tilde{\vect{x}}_t = \vect{x}^\text{sim}_t$, while the privileged observation passed to the encoder during training additionally contains the future reference hand kinematics, $\vect{x}_t = (\tilde{\vect{x}}_t, \vect{x}^\text{ref}_t)$. Both the prior and the decoder are conditioned only on $\tilde{\vect{x}}_t$. The distillation loss and KL regularization are identical to our interaction prior.

\subsubsection{Latent Exploration}
For the generative tasks, we add the table and the object to the scene. The exploration policy $\pi_\text{explore}$ therefore needs to know the object and its target trajectory, so it receives the object-aware partial observation $\tilde{\vect{o}}_t$ of our main method rather than the restricted $\tilde{\vect{x}}_t$, and predicts a residual $\Delta \vect{z}_t = \pi_\text{explore}(\tilde{\vect{o}}_t)$. Because the pre-trained decoder remains blind to the object, this residual is added to the motion-only prior mean and decoded from the proprioceptive state alone, $\vect{a}_t = \mathcal{D} \big( \tilde{\vect{x}}_t, \mu_\mathcal{P}(\tilde{\vect{x}}_t) + \Delta \vect{z}_t \big)$. This restricts exploration to human-like motions but deprives the decoder of any explicit interaction awareness.

\subsection{Additional Rewards and Curriculum for Tabletop Manipulation}
\label{app:additional_rewards_curriculum}

For Tabletop Manipulation, our method is trained with the object surface-sample reward alone and directly on the full task distribution. The baselines have no prior exposure to objects, and with this setup the initially uninformative tracking signal rarely leads the hands to reach the object or establish a stable grasp; in our experiments, neither baseline learned meaningful behavior within 6\,h of training. To accelerate their learning, we add a pre-grasp reward and a training curriculum, as summarized in \Cref{tab:task_rewards_curriculum}. Apart from these two additions, the baselines follow the training procedure of \Cref{subsec:generative_tasks_v2}.

\begin{table}[h]
    \centering
    \caption{Reward terms and curriculum used by each method for Tabletop Manipulation. Although our method relies only on the object surface-sample reward and needs no curriculum, it outperforms both baselines.}
    \label{tab:task_rewards_curriculum}
    \small
    \renewcommand{\arraystretch}{1.1}
    \begin{tabular}{l ccc}
        \toprule
        Method & Obj.\ Sample & Pre-grasp & Curriculum \\
        \midrule
        Action Space Exploration & \checkmark & \checkmark & \checkmark \\
        Hand-Only Prior          & \checkmark & \checkmark & \checkmark \\
        Ours                     & \checkmark & $\times$   & $\times$   \\
        \bottomrule
    \end{tabular}
\end{table}

\noindent\textbf{Pre-grasp reward.}
Following~\cite{dasari2023learning, luo2024ominigrasp}, we add a pre-grasp reward term. Since each episode initializes from a known training-pool state, we have access to the corresponding ground-truth pre-grasp hand pose $\hat{\vect{q}}^\text{pg}$ at the first contact frame. We use it \emph{only} as a reward signal and never as a policy input, which preserves the zero-shot setting at inference. The pre-grasp loss $L^\text{pregrasp}$ is the Euclidean keypoint distance between the simulated hands and $\hat{\vect{q}}^\text{pg}$. We split each episode into two sequential phases. During the initial approach phase, the reward only encourages reaching the pre-grasp pose,
\begin{equation*}
r_t = \exp\!\big(-\lambda_\text{grasp}\,L^\text{pregrasp}\big),
\end{equation*}
and after a fixed number of initial steps (40 frames), it switches to the object tracking objective,
\begin{equation*}
r_t = \exp\!\big(-\lambda_\text{samp}\,L^\text{samp}\big).
\end{equation*}
This explicit transition prevents the two objectives from interfering, so the policy first focuses on establishing a stable force closure before shifting its capacity to realizing the target trajectory.

\noindent\textbf{Curriculum.}
Direct training over the highly diverse initial poses and trajectories proves difficult for the baselines, so we introduce a three-level curriculum that progressively increases task complexity. In level~1, environments are initialized with matched hand and object poses drawn from the same training sequence. Level~2 adds continuous $z$-axis rotation augmentations to both the initial poses and the target trajectory. Level~3 completely decouples the initial configurations of hands and object by sampling the hand pose from a different training sequence than the object pose. Our method is trained on this highest level from the start.

\section{Experiment Details and Additional Results}
\label{app:experiments}

\subsection{Comparison with ManipTrans}
\label{sec:appendix_maniptrans_comp}

\begin{figure}[h]
    \centering
    \includegraphics[width=0.5\linewidth]{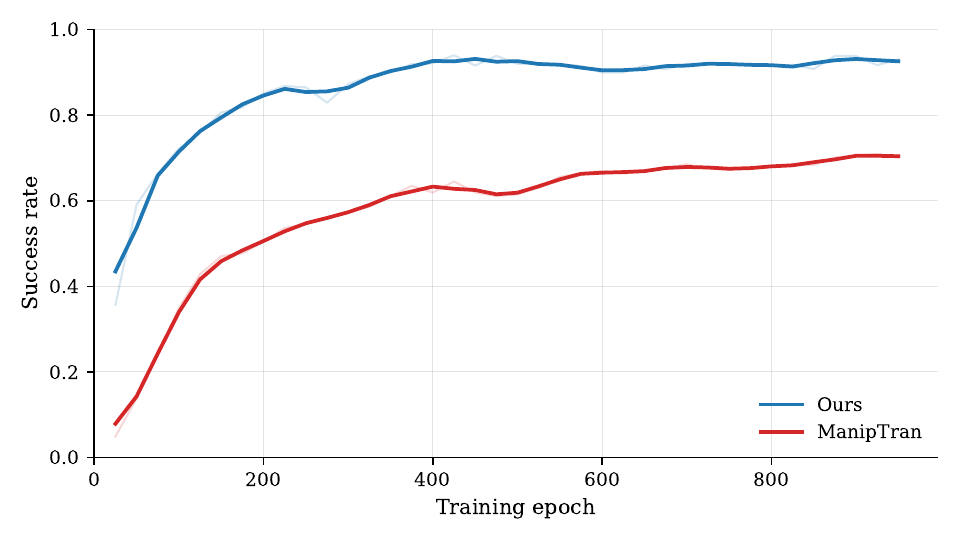}
    \caption{Success rate over training epochs of our imitation experts compared to ManipTrans on GRAB50.}
    \label{fig:maniptrans_training_curves}
\end{figure}
To evaluate our expert training against ManipTrans~\citep{li2025maniptrans}, we sample a subset of 50 right-handed sequences from the GRAB dataset. Because ManipTrans relies on a distinct, predefined MANO-based robotic hand model, we first retarget the source mocap data to their embodiment to ensure that their tracking controllers function correctly. Our formulation, in contrast, absorbs these morphological discrepancies and does not need explicit kinematic retargeting.

Both methods are trained with identical hyperparameters for 1,000 epochs to ensure convergence, and evaluated with a strict 5\,cm object translation threshold. The only difference to the per-object hyperparameters in \Cref{tab:expert_hyperparameters} is the reduced minibatch size of 1024. \Cref{tab:eval-expert-full} reports the results over three seeds. \Cref{fig:maniptrans_training_curves} further shows that our method is considerably more sample efficient: it reaches the peak success rate of ManipTrans after only 175 epochs and ultimately converges to a clearly higher success rate and spatial precision.

\begin{table}[t]
    \centering
    \caption{Three-seed replication of the ManipTrans comparison on GRAB50. We report mean $\pm$ standard deviation over 3 random seeds.}
    \label{tab:eval-expert-full}
    \small
    \setlength{\tabcolsep}{5pt}
    \renewcommand{\arraystretch}{1.1}
    \resizebox{\linewidth}{!}{%
    \begin{tabular}{l l l c c c c c}
    \toprule
    Method & Data & Granularity & SR (5cm) [\%] & TC (5cm) [\%] & $E_t$ [cm] & $E_r$ [$^\circ$] & MPJPE [cm] \\
    \midrule
    ManipTrans~\citep{li2025maniptrans}
        & GRAB50 & Per-sequence & $69.6 \pm 1.5$ & $77.6 \pm 1.9$ & $0.43 \pm 0.01$ & $7.3 \pm 0.2$ & $1.97 \pm 0.06$ \\
    Ours
        & GRAB50 & Per-sequence & $\mathbf{93.3 \pm 3.0}$ & $\mathbf{95.8 \pm 2.6}$ & $\mathbf{0.28 \pm 0.05}$ & $\mathbf{4.9 \pm 1.2}$ & $\mathbf{1.01 \pm 0.10}$ \\
    \bottomrule
    \end{tabular}
    }
\end{table}

\subsection{Tabletop Manipulation}
\label{app:full_generative_results}

\Cref{tab:generative_evaluation_full} reports the Tabletop Manipulation results on GRAB as mean and standard deviation over three random seeds. The gain over the baselines is consistent across seeds, and our method also shows by far the lowest variance.

\begin{table}[h]
    \centering
    \caption{Tabletop Manipulation on held-out GRAB objects. We report mean $\pm$ standard deviation over 3 random seeds. SR and TC in \% (10\,cm threshold), $E_t$ in cm, $E_r$ in degrees.}
    \label{tab:generative_evaluation_full}
    \small
    \setlength{\tabcolsep}{4pt}
    \renewcommand{\arraystretch}{1.1}
    \begin{tabular}{l cccc}
        \toprule
        Method & SR $\uparrow$ & TC $\uparrow$ & $E_t$ $\downarrow$ & $E_r$ $\downarrow$ \\
        \midrule
        Action Space Exploration & $59.1 \pm 4.7$ & $73.6 \pm 3.6$ & $1.79 \pm 0.10$ & $21.5 \pm 1.0$ \\
        Hand-Only Prior          & $60.0 \pm 3.1$ & $73.4 \pm 2.1$ & $1.52 \pm 0.06$ & $21.2 \pm 1.0$ \\
        Ours                     & $\mathbf{74.0 \pm 0.9}$ & $\mathbf{82.9 \pm 0.6}$ & $\mathbf{1.36 \pm 0.01}$ & $\mathbf{20.0 \pm 0.5}$ \\
        \bottomrule
    \end{tabular}
\end{table}

\subsection{Novel Tasks: Common Setup}
\label{app:downstream}
\label{app:downstream_common}

Both novel tasks of catching and bimanual handover reuse the generative machinery of \cref{sec:experiments}: a task is specified entirely by an object reference trajectory, and $\pi_\text{explore}$ is optimized with an object-tracking reward, complemented by a small number of task-specific terms described below. This section defines the shared setup, and the following two sections define the reference trajectories, task-specific terminations, and the training and evaluation protocol of each task.

\noindent\textbf{Scene.}
Neither task involves a physical table. The bimanual hands float in free space.

\noindent\textbf{Reward.}
The reward is the exponentiated combination of an object position term (weight 10) and a fingertip-proximity term (weight 10 for catching, 5 for handover).

\noindent\textbf{Termination.}
An episode terminates when the object leaves a task-specific radius around the reference (30\,cm for catching, 15\,cm for handover), when the nearest wrist is farther than a task-specific distance from the object (2.0\,m and 1.2\,m), or on self-penetration.

\noindent\textbf{Objects.}
Catching uses all 45 training objects and the 5 held-out objects of \cref{sec:experiments}. Handover requires a left-hand grasp frame to initialize from; five training objects (\texttt{knife}, \texttt{scissors}, \texttt{teapot}, \texttt{toothbrush}, \texttt{torussmall}) have none in GRAB and are excluded, leaving 40. All 5 held-out objects have left-hand grasp frames in the evaluation split.

\subsection{Catching}
\label{app:downstream_catch}

\noindent\textbf{Reference trajectory.}
The object is launched on a ballistic arc that passes through an intercept point, and the reference then holds the object at that point for at least 60 frames. To make the reference physically exact, the arc is integrated with the simulator's own symplectic Euler scheme rather than the closed-form parabola. After $k$ substeps the position is
\begin{equation*}
\mathbf{p}_k = \mathbf{p}_0 + k\,\Delta t\,\mathbf{v}_0 + \mathbf{g}\,\Delta t^2\,k(k{+}1)/2,
\end{equation*}
and the launch velocity is solved from this expression. Objects are given a random initial orientation and a random angular velocity of up to 8\,rad/s; because the simulator rotates rigid bodies about the center of mass, the reference is computed for the center of mass and mapped back to the object origin using the per-object inertial offset. Episodes last 160 frames (2.67\,s at 60\,Hz).

\noindent\textbf{Randomization.}
The intercept point is displaced from the nominal catch point by up to $(\pm0.35, \pm0.25, \pm0.30)$\,m, the hands start at a common random offset of up to $(\pm0.15, \pm0.10, \pm0.10)$\,m, the launch position lies 0.3--1.0\,m from the intercept point at an azimuth of $\pm75^\circ$ and a height of $-0.3$ to $+0.6$\,m relative to it, and flight time is 15--70 frames (0.25--1.17\,s). Both hands start open, palms facing each other 12\,cm either side of the nominal catch point.

\noindent\textbf{Wrist economy.}
With the tracking reward alone, interaction prior-based policies learn to send one hand out to the object early and ride along with it for the entire flight, a behavior the prior's data encourages (a GRAB hand is always near the object) and that the fingertip-proximity term pays for on every flight frame. We add two penalties on the wrists' linear and angular speed (weights 2.0 and 0.3), applied only before the object reaches the intercept point, and scale the fingertip-proximity term to 10\% of its weight during the same window.

\noindent\textbf{Right-hand catching.}
In this variant only the right hand may touch the object. Contact between the left hand and the object terminates the episode, and the fingertip-proximity and wrist-economy terms are computed for the right hand only. Because a single hand covers a smaller workspace, throws are narrower and flatter. The launch position lies 0.3--0.8\,m from the intercept point at an azimuth of $\pm50^\circ$ and a height of $-0.2$ to $+0.45$\,m relative to it, and the intercept point is displaced from the nominal catch point by up to $(\pm0.25, \pm0.18, \pm0.22)$\,m. All other settings are identical to bimanual catching.

\noindent\textbf{Training and evaluation.}
All policies are trained for 20h on a single A100. All methods receive the same wall-clock budget; because action space exploration does not evaluate a decoder at every step, it completes more updates in this time. Policies are evaluated deterministically with 20 episodes per object, giving 900 episodes on the 45 training objects and 100 on the 5 held-out objects. Throws are generated from a fixed seed, so every method is evaluated on the same sequence of objects, launch states and intercept points.

\noindent\textbf{Results.}
\Cref{tab:catch_evaluation_full} reports the full comparison. On the held-out objects, our bimanual policy catches the object in 87.0\% of the episodes, compared with 43.0\% for action space exploration, while the hand-only prior almost never succeeds. The picture is the same on the training objects, and our policy performs as well on the held-out objects as on the training objects. Restricting the catch to the right hand lowers the success rate to 78.0\%, but even this harder variant clearly outperforms both bimanual baselines.

\begin{table}[h]
    \centering
    \caption{Catching a thrown object. \textit{Left:} the 45 training objects, 900 episodes per method. \textit{Right:} five held-out objects, 100 episodes per method. All methods are trained under an equal wall-clock budget. The \emph{right only} row is a different task, catching with the right hand only (\cref{app:downstream_catch}), and is not directly comparable to the other rows.}
    \label{tab:catch_evaluation_full}
    \small
    \setlength{\tabcolsep}{4pt}
    \renewcommand{\arraystretch}{1.1}
    \begin{tabular}{l l ccc @{\hspace{3ex}} ccc}
        \toprule
        & & \multicolumn{3}{c}{Training Objects} & \multicolumn{3}{c}{Held-Out Objects} \\
        \cmidrule(r{3ex}){3-5} \cmidrule(l){6-8}
        Method & Hand Usage & SR $\uparrow$ & TC $\uparrow$ & $E_t$ $\downarrow$ & SR $\uparrow$ & TC $\uparrow$ & $E_t$ $\downarrow$ \\
        \midrule
        Action Space Exploration & both hands & 51.1 & 68.2 & 3.19 & 43.0 & 61.9 & 3.27 \\
        Hand-Only Prior          & both hands & 4.9  & 37.3 & 3.70 & 1.0  & 32.3 & 3.55 \\
        Ours                     & right only & 78.2 & 89.8 & 2.03 & 78.0 & 88.3 & 1.98 \\
        Ours                     & both hands & \textbf{84.6} & \textbf{91.6} & \textbf{1.66} & \textbf{87.0} & \textbf{94.5} & \textbf{1.38} \\
        \bottomrule
    \end{tabular}
\end{table}

\subsection{Handover}
\label{app:downstream_handover}

\noindent\textbf{Initial state.}
Each episode starts from a GRAB frame in which the left hand alone holds the lifted object: the object center is more than 10\,cm above the table, at least four left-hand contact bodies and at most one right-hand contact body touch it, and its speed is below 0.2\,m/s. Where the object has same-actor handover sequences (GRAB intent \emph{offhand}), 70\% of resets are drawn from the pre-transfer part of those sequences. The hands and object are translated so that the object starts on the left side of the workspace, $x\in[0.30, 0.45]$\,m in the table frame, with the right hand at its GRAB pose relative to the object. Five training objects (\texttt{knife}, \texttt{scissors}, \texttt{teapot}, \texttt{toothbrush}, \texttt{torussmall}) have no such frame and are excluded, leaving 22{,}641 start frames over 40 training objects. The five held-out objects have 1{,}442 start frames in the evaluation split.

A GRAB frame copied into the simulator is kinematically correct but not necessarily force-closed: the object may slip out of the left hand before any policy acts. To separate the handover from the problem of establishing a stable grasp in the first place, we test every held-out start frame without a policy. The hands are held at their initial joint configuration and wrist pose by the PD controller, and we record the displacement of the object from its initial position. A frame is kept if the object moves less than 5\,cm within the first 30 frames (0.5\,s). Because the test involves no policy, it selects the same states for every method. Of the 1{,}442 held-out frames, 160 pass.

\noindent\textbf{Reference trajectory.}
The reference holds the object in place for 30 frames (\emph{settle}), during which terminations are masked so that the policy can establish its own grip. It then moves the object in a straight line at constant height to $x\in[-0.45, -0.30]$\,m at 0.2--0.4\,m/s, with 0.5\,s velocity ramps and a 0.5\,s dwell at the midpoint, and finally holds it for at least 60 frames. Episodes last 420 frames (7\,s).

\noindent\textbf{Wrist zones.}
The left wrist may not move right of $x=-0.08$\,m and the right wrist may not move left of $x=+0.08$\,m. The zones are enforced by clamping the wrist position targets, and an episode terminates if a wrist is pushed more than 2\,cm past its zone. Since both zones exclude the far side of the workspace, no single hand can carry the object from start to goal: an episode that runs to completion has brought the object into the right hand's zone. 

\noindent\textbf{Per-hand proximity.}
The fingertip-proximity term takes the nearest fingertip of each hand separately and sums the two distances, so that the free hand receives a gradient towards the object instead of being masked by the hand already holding it.

\noindent\textbf{Metrics.}
SR and TC are defined as before. The translation error $E_t$ is the mean distance between object and reference from the end of the settle phase until termination.

\noindent\textbf{Evaluation.}
On the training objects we run 800 episodes per method from randomly drawn start frames. On the held-out objects we run one episode per stable start frame and method (160 episodes). All methods see identical start states, and policies are evaluated deterministically. Full results are in \Cref{tab:handover_evaluation_full}.

\begin{table}[h]
    \centering
    \caption{Bimanual handover. \textit{Left:} training objects, 800 episodes per method. \textit{Right:} held-out objects, restricted to the 160 motion-capture start frames that are physically stable (\cref{app:downstream_handover}). All methods start from identical states.}
    \label{tab:handover_evaluation_full}
    \label{tab:app_handover_transfer}
    \small
    \setlength{\tabcolsep}{3.5pt}
    \renewcommand{\arraystretch}{1.1}
    \begin{tabular}{l ccc @{\hspace{3ex}} ccc}
        \toprule
        & \multicolumn{3}{c}{Training Objects} & \multicolumn{3}{c}{Held-Out Objects} \\
        \cmidrule(r{3ex}){2-4} \cmidrule(l){5-7}
        Method & SR $\uparrow$ & TC $\uparrow$ & $E_t$ $\downarrow$ & SR $\uparrow$ & TC $\uparrow$ & $E_t$ $\downarrow$ \\
        \midrule
        Action Space Exploration & 53.8 & 69.2 & 2.46 & 11.9 & 30.8 & 5.27 \\
        Hand-Only Prior          & 26.8 & 51.2 & 4.39 &  5.0 &  35.3 & 4.45 \\
        Ours                     & \textbf{69.0} & \textbf{79.4} & \textbf{1.60} & \textbf{51.9} & \textbf{74.6} & \textbf{2.18} \\
        \bottomrule
    \end{tabular}
\end{table}

\section{Robustness and Ablations}
\label{app:robustness_ablations}

We run all experiments in this section on Tabletop Manipulation on GRAB.

\subsection{Robustness to Physical Perturbations}
\label{app:robustness}

We test how the Tabletop Manipulation policies of all three methods cope with physical conditions that differ from training, varying one factor at a time on the held-out objects (\Cref{tab:robustness}). Changing the surface friction affects all methods similarly, and higher friction makes grasping easier for every method. Heavier objects, however, clearly separate the methods: at four times the nominal mass, our policy loses 11.7\% of its success rate, while the hand-only prior and action space exploration lose 67.2\% and 52.8\%. Our policy is also the most robust to action noise. Under all perturbations our method retains the highest absolute success rate.

\begin{table}[h]
\caption{Zero-shot robustness of the Tabletop Manipulation policies to physical perturbations. Success rate in \% on held-out objects (100 evaluations per motion), varying one factor at a time; parentheses give the relative change w.r.t.\ each method's own base setting (friction 0.7, mass $1\times$, no noise). Action noise adds zero-mean Gaussian noise to the commanded joint targets at every control step (standard deviation in degrees for rotational joints and in mm for wrist translation). Single seed only.}
  \label{tab:robustness}
  \centering
  \small
  \setlength{\tabcolsep}{5pt}
  \newcommand{\dlt}[1]{{\scriptsize(#1)}}

  \begin{subtable}{\linewidth}
    \centering
    \caption{Surface friction}
    \label{tab:robust_friction}
    \begin{tabular}{lccc}
      \toprule
      Friction & Action Space Exploration & Hand-Only Prior & Ours \\
      \midrule
      0.35 (slippery)     & 53.8 \dlt{$-$8.0\%}  & 58.5 \dlt{$-$6.8\%}  & 68.5 \dlt{$-$5.4\%} \\
      0.7 \textit{(base)} & 58.5                 & 62.8                 & 72.4                \\
      1.4                 & 63.4 \dlt{$+$8.4\%}  & 67.2 \dlt{$+$7.0\%}  & 76.4 \dlt{$+$5.5\%} \\
      2.1 (grippy)        & 66.1 \dlt{$+$13.0\%} & 69.0 \dlt{$+$9.9\%}  & 78.2 \dlt{$+$8.0\%} \\
      \bottomrule
    \end{tabular}
  \end{subtable}

  \vspace{0.6em}

  \begin{subtable}{\linewidth}
    \centering
    \caption{Object mass multiplier}
    \label{tab:robust_mass}
    \begin{tabular}{lccc}
      \toprule
      Mass & Action Space Exploration & Hand-Only Prior & Ours \\
      \midrule
      0.5$\times$ (light)       & 59.7 \dlt{$+$2.1\%}  & 63.9 \dlt{$+$1.8\%}  & 72.1 \dlt{$-$0.5\%}  \\
      1$\times$ \textit{(base)} & 58.5                 & 62.8                 & 72.4                 \\
      2$\times$                 & 48.2 \dlt{$-$17.6\%} & 40.3 \dlt{$-$35.8\%} & 71.4 \dlt{$-$1.4\%}  \\
      4$\times$ (heavy)         & 27.6 \dlt{$-$52.8\%} & 20.6 \dlt{$-$67.2\%} & 64.0 \dlt{$-$11.7\%} \\
      \bottomrule
    \end{tabular}
  \end{subtable}

  \vspace{0.6em}

  \begin{subtable}{\linewidth}
    \centering
    \caption{Action noise}
    \label{tab:robust_actnoise}
    \begin{tabular}{lccc}
      \toprule
      Noise & Action Space Exploration & Hand-Only Prior & Ours \\
      \midrule
      0 \textit{(base)}  & 58.5                 & 62.8                 & 72.4                 \\
      1$^\circ$ / 1\,mm  & 54.8 \dlt{$-$6.3\%}  & 52.6 \dlt{$-$16.2\%} & 67.6 \dlt{$-$6.6\%}  \\
      2$^\circ$ / 2\,mm  & 46.2 \dlt{$-$21.0\%} & 32.4 \dlt{$-$48.4\%} & 60.7 \dlt{$-$16.2\%} \\
      4$^\circ$ / 4\,mm  & 27.0 \dlt{$-$53.8\%} & 14.6 \dlt{$-$76.8\%} & 42.6 \dlt{$-$41.2\%} \\
      \bottomrule
    \end{tabular}
  \end{subtable}
\end{table}

\subsection{Sensitivity to the Termination Threshold}
\label{app:threshold}

Tabletop Manipulation uses a relaxed termination threshold of 10\,cm (\Cref{subsec:tabletop}). To verify that our results do not depend on this choice, we evaluate the same policies with stricter object-position thresholds of 7.5\,cm and 5\,cm (\Cref{tab:threshold_sensitivity}). All methods lose success as the threshold tightens, but our method degrades the least: from 10\,cm to 5\,cm, its success rate drops by 10.5 points, compared with 13.4 points for the hand-only prior and 21.9 points for action space exploration. The gap to the strongest baseline therefore grows from 14.0 to 16.9 points, which indicates that our policies track the target trajectory more closely rather than merely staying within the tolerance.

\begin{table}[h]
  \centering
  \small
  \setlength{\tabcolsep}{8pt}
  \renewcommand{\arraystretch}{1.1}
  \caption{Tabletop Manipulation on held-out GRAB objects under stricter object-position termination thresholds. Success rate in \%, mean $\pm$ standard deviation over 3 seeds.}
  \label{tab:threshold_sensitivity}
  \begin{tabular}{l ccc}
    \toprule
    Method & SR @ 10\,cm $\uparrow$ & SR @ 7.5\,cm $\uparrow$ & SR @ 5\,cm $\uparrow$ \\
    \midrule
    Action Space Exploration & 59.1 $\pm$ 4.7 & 50.0 $\pm$ 4.0 & 37.2 $\pm$ 4.0 \\
    Hand-Only Prior          & 60.0 $\pm$ 3.1 & 55.0 $\pm$ 2.6 & 46.6 $\pm$ 2.4 \\
    Ours                     & \textbf{74.0 $\pm$ 0.9} & \textbf{69.9 $\pm$ 1.1} & \textbf{63.5 $\pm$ 1.1} \\
    \bottomrule
  \end{tabular}
\end{table}

\subsection{Stage~III Ablations}
\label{app:ablations}

\Cref{tab:ablations_full} compares our Stage~III policy with three ablated variants. The two trained variants use the same observations, reward and PPO settings as our full policy.

\noindent\textbf{Direct prior rollout.}
This variant executes the frozen prior and decoder without any exploration policy, $\vect{a}_t=\mathcal{D}\big(\vect{o}^\text{sim}_t,\mu_\mathcal{P}(\tilde{\vect{o}}_t)\big)$, and thus measures how much Stage~III adds on top of the distilled prior.

\noindent\textbf{Action-space residual.}
This variant keeps the frozen prior and decoder, but the exploration policy adds its residual to the decoded action instead of to the latent, $\vect{a}_t=\mathcal{D}\big(\vect{o}^\text{sim}_t,\mu_\mathcal{P}(\tilde{\vect{o}}_t)\big)+\Delta\vect{a}_t$.

\noindent\textbf{W/o random trajectories.}
This variant trains Stage~III only on the recorded GRAB object trajectories, while keeping all other augmentations such as the randomized initial states.

\begin{table}[h!]
  \centering
  \small
  \setlength{\tabcolsep}{8pt}
  \renewcommand{\arraystretch}{1.1}
  \caption{Stage~III ablations on held-out GRAB objects. $E_t$ in cm, $E_r$ in degrees.}
  \label{tab:ablations_full}
  \begin{tabular}{l cccc}
    \toprule
    Method & SR $\uparrow$ & TC $\uparrow$ & $E_t$ $\downarrow$ & $E_r$ $\downarrow$ \\
    \midrule
    Direct prior rollout    & 30.8 & 50.4 & 2.19 & 20.8 \\
    Action-space residual   & 49.6 & 65.2 & 1.87 & 21.0 \\
    W/o random traj.        & 67.9 & 78.4 & 1.37 & 20.4 \\
    \midrule
    Ours                    & \textbf{74.0} & \textbf{82.9} & \textbf{1.36} & \textbf{20.0} \\
    \bottomrule
  \end{tabular}
\end{table}

\subsection{Residual Bound}
\label{app:bound}

We ablate the residual bound $B$ by training Stage~III policies with $B \in \{0.5, 1, 2, 3, 5, 10, \infty\}$ and evaluating them on held-out objects at matched training epochs (\Cref{fig:bound_ablation}). The bound trades task success against grasp diversity. Tight bounds leave the policy too little room to correct the prior, and success rises steadily as the bound is relaxed (\Cref{fig:bound_ablation}\subref{fig:bound_sr}). Up to $B{=}5$, the deployed latents stay within the prior's latent distribution (\Cref{fig:bound_ablation}\subref{fig:bound_pca}), and hand usage closely follows the human reference (\Cref{fig:bound_ablation}\subref{fig:bound_hands}). For $B \geq 10$ the bound no longer constrains the residual, and $B{=}10$ behaves like the unbounded policy. Success improves only marginally, but the latents drift away from the prior and the policy collapses almost entirely to bimanual grasps, discarding the grasp diversity encoded in the prior. We therefore use $B{=}5$, which retains most of the unbounded success rate while keeping the hand usage close to the human reference.

\end{document}